# Generating Natural Questions About an Image

**Nasrin Mostafazadeh[1], Ishan Misra[2], Jacob Devlin[3], Margaret Mitchell[3]**
**Xiaodong He[3], Lucy Vanderwende[3]**
1 University of Rochester, 2 Carnegie Mellon University,
3 Microsoft Research
nasrinm@cs.rochester.edu, lucyv@microsoft.com

## Abstract

There has been an explosion of work in the vision & language community during the past few years from image captioning to video transcription, and answering questions about images. These tasks have focused on literal descriptions of the image. To move beyond the literal, we choose to explore how questions about an image are often directed at commonsense inference and the abstract events evoked by objects in the image. In this paper, we introduce the novel task of **Visual Question Generation (VQG)**, where the system is tasked with asking a natural and engaging question when shown an image. We provide three datasets which cover a variety of images from object-centric to event-centric, with considerably more abstract training data than provided to state-of-the-art captioning systems thus far. We train and test several generative and retrieval models to tackle the task of VQG. Evaluation results show that while such models ask reasonable questions for a variety of images, there is still a wide gap with human performance which motivates further work on connecting images with commonsense knowledge and pragmatics. Our proposed task offers a new challenge to the community which we hope furthers interest in exploring deeper connections between vision & language.

## 1 Introduction

We are witnessing a renewed interest in interdisciplinary AI research in vision & language, from descriptions of the visual input such as image captioning (Chen et al., 2015; Fang et al., 2014; Donahue et al., 2014; Chen et al., 2015) and video transcription (Rohrbach et al., 2012; Venugopalan et al., 2015), to testing computer understanding of an image through question answering (Antol et al., 2015; Malinowski and Fritz, 2014). The most established work in the vision & language community is 'image captioning', where the task is to produce a literal description of the image. It has been shown (Devlin et al., 2015; Fang et al., 2014; Donahue et al., 2014) that a reasonable language modeling paired with deep visual features trained on large enough datasets promise a good performance on image captioning, making it a less challenging task from language learning perspective. Furthermore, although this task has a great value for communities of people who are low-sighted or cannot see in all or some environments, for others, the description does not add anything to what a person has already perceived.

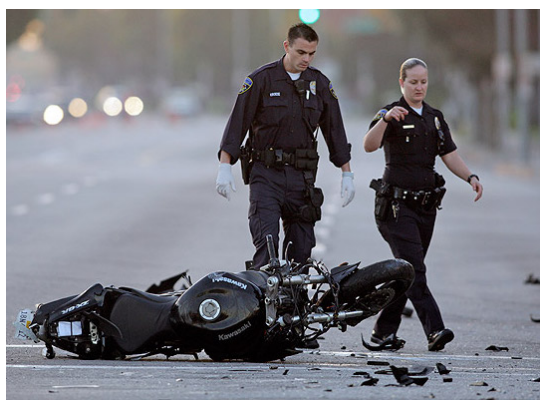

**Natural Questions:**
- Was anyone injured in the crash?
- Is the motorcyclist alive?
- What caused this accident?

**Generated Caption:**
- A man standing next to a motorcycle.

Figure 1: Example image along with its natural questions and automatically generated caption.

The popularity of the image sharing applications in social media and user engagement around images is evidence that commenting on pictures is a very natural task. A person might respond to an image with a short comment such as 'cool', 'nice pic' or ask a question. Imagine someone has shared the image in Figure 1. What is the very first question that comes to mind? Your question is most probably very similar to the questions listed next to the image, expressing concern about the motorcyclist (who is not even present in the image). As you can tell, natural questions are not about what is seen, the policemen or the motorcycle, but rather about what is inferred given these objects, e.g., an accident or injury. As such, questions are often about abstract concepts, i.e., events or states, in contrast to the concrete terms[1] used in image captioning. It is clear that the corresponding automatically generated caption[2] for Figure 1 presents only a literal description of objects.

To move beyond the literal description of image content, we introduce the novel task of Visual Question Generation (VQG), where given an image, the system should 'ask a natural and engaging question'. Asking a question that can be answered simply by looking at the image would be of interest to the Computer Vision community, but such questions are neither natural nor engaging for a person to answer and so are not of interest for the task of VQG.

Learning to ask questions is an important task in NLP and is more than a syntactic transformation of a declarative sentence (Vanderwende, 2008). Deciding what to ask about demonstrates understanding and as such, question generation provides an indication of machine understanding, just as some educational methods assess students' understanding by their ability to ask relevant questions[3]. Furthermore, training a system to ask a good question (not only answer a question) may imbue the system with what appears to be a cognitive ability unique to humans among other primates (Jordania, 2006). Developing the ability to ask relevant and to-the-point questions can be an essential component of any dynamic learner which seeks information. Such an ability can be an integral component of any conversational agent, either to engage the user in starting a conversation or to elicit task-specific information.

The contributions of this paper can be summarized as follows: (1) in order to enable the VQG research, we carefully created three datasets with a total of 75,000 questions, which range from object- to event-centric images, where we show that VQG covers a wide range of abstract terms including events and states (Section 3). (2) we collected 25,000 gold captions for our event-centric dataset and show that this dataset presents challenges to the state-of-the-art image captioning models (Section 3.3). (3) we perform analysis of various generative and retrieval approaches and conclude that end-to-end deep neural models outperform other approaches on our most-challenging dataset (Section 4). (4) we provide a systematic evaluation methodology for this task, where we show that the automatic metric $\Delta$BLEU strongly correlates with human judgments (Section 5.3). The results show that while our models learn to generate promising questions, there is still a large gap to match human performance, making the generation of relevant and natural questions an interesting and promising new challenge to the community.

## 2 Related Work

For the task of image captioning, datasets have primarily focused on objects, e.g. Pascal VOC (Everingham et al., 2010) and Microsoft Common Objects in Context (MS COCO) (Lin et al., 2014). MS COCO, for example, includes complex everyday scenes with 91 basic objects in 328k images, each paired with 5 captions. Event detection is the focus in video processing and action detection, but these do not include a textual description of the event (Yao et al., 2011b; Andriluka et al., 2014; Chao et al., 2015; Xiong et al., 2015). The number of actions in each of these datasets is still relatively small, ranging from 40 (Yao et al., 2011a) to 600 (Chao et al., 2015) and all involve human-oriented activity (e.g. 'cooking', 'gardening', 'riding a bike'). In our work, we are focused on generating questions for static images of events, such as 'fire', 'explosion' or 'snowing', which have not yet been investigated in any of the above datasets.

Visual Question Answering is a relatively new task where the system provides an answer to a question about the image content. The most notable, Visual Question Answering (VQA) (Antol et al., 2015), is an open-ended (free-form) dataset, in which both the questions and the answers are crowd-sourced, with workers prompted to ask a visually verifiable question which will 'stump a smart robot'. Gao et al. (2015) used similar methodology to create a visual question answering dataset in Chinese. COCO-QA (CQA) (Ren et al., 2015), in contrast, does not use human-authored questions, but generates questions automatically from image captions of the MS COCO dataset by applying a set of transformation rules to generate the wh-question. The expected answers in CQA

[1] Concrete terms are the ones that can be experienced with five senses. Abstract terms refer to intangible things, such as feelings, concepts, and qualities

[2] Throughout this paper we use the state-of-the-art captioning system (Fang et al., 2014), henceforth MSR captioning system `https://www.captionbot.ai/`, to generate captions.

[3] `http://rightquestion.org/`

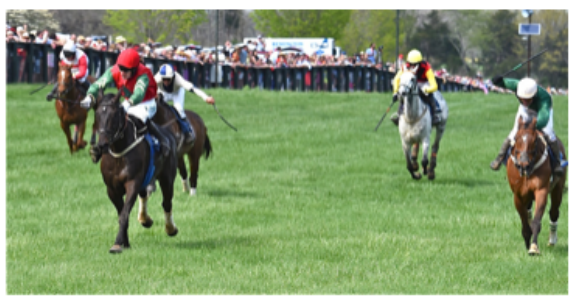

- How many horses are in the field? ✗

- Who won the race? ✓

Figure 2: Example right and wrong questions for the task of VQG.

are by design limited to objects, numbers, colors, or locations. A more in- depth analysis of VQA and CQA datasets will be presented in Section 3.1.

In this work, we focus on questions which are interesting for a person to answer, not questions designed to evaluate computer vision. A recently published work on VQA, Visual7W (Zhu et al., 2016), establishes a grounding link on the object regions corresponding to the textual answer. This setup enables a system to answer a question with visual answers (in addition to textual answers). They collect a set of 327,939 7W multiple-choice QA pairs, where they point out that 'where', 'when' and 'why' questions often require high-level commonsense reasoning, going beyond spatial reasoning required for 'which' or 'who' questions. This is more in line with the type of questions that VQG captures, however, the majority of the questions in Visual7w are designed to be answerable by only the image, making them unnatural for asking a human. Thus, learning to generate the questions in VQA task is not a useful sub-task, as the intersection between VQG and any VQA questions is by definition minimal.

Previous work on question generation from textual input has focused on two aspects: the grammaticality (Wolfe, 1976; Mitkov and Ha, 2003; Heilman and Smith, 2010) and the content focus of question generation, i.e., "what to ask about". For the latter, several methods have been explored: (Becker et al., 2012) create fill-in-the-blank questions, (Mazidi and Nielsen, 2014) and (Lindberg et al., 2013) use manually constructed question templates, while (Labutov et al., 2015) use crowdsourcing to collect a set of templates and then rank the potentially relevant templates for the selected content. To our knowledge, neither a retrieval model nor a deep representation of textual input, presented in our work, have yet been used to generate questions.

## 3 Data Collection Methodology

**Task Definition:** Given an image, the task is to generate a natural question which can potentially engage a human in starting a conversation. Questions that are visually verifiable, i.e., that can be answered by looking at only the image, are outside the scope of this task. For instance, in Figure 2, a question about the number of horses (appearing in the VQA dataset) or the color of the field is not of interest. Although in this paper we focus on asking a question about an image in isolation, adding prior context or history of conversation is the natural next step in this project.

We collected the VQG questions by crowdsourcing the task on Amazon Mechanical Turk (AMT). We provide details on the prompt and the specific instructions for all the crowdsourcing tasks in this paper in the supplementary material. Our prompt was very successful at capturing nonliteral questions, as the good question in Figure 2 demonstrates. In the following Sections, we describe our process for selecting the images to be included in the VQG dataset. We start with images from MS COCO, which enables meaningful comparison with VQA and CQA questions. Given that it is more natural for people to ask questions about event-centric images, we explore sourcing eventful images from Flickr and from querying an image search engine. Each data source is represented by 5,000 images, with 5 questions per image.

### 3.1 $VQG_{coco-5000}$ and $VQG_{Flickr-5000}$

As our first dataset, we collected VQG questions for a sample of images from the MS COCO dataset[4]. In order to enable comparisons with related datasets, we sampled 5,000 images of MS COCO which were also annotated by the CQA dataset (Ren et al., 2015) and by VQA (Antol et al., 2015). We name this dataset $VQG_{coco-5000}$. Table 1 shows a sample MS COCO image along with annotations in the various datasets. As the CQA questions are generated by rule application from captions, they are not always coherent. The VQA questions are written to evaluate the detailed visual understanding of a robot, so their questions are mainly visually grounded and literal. The table demonstrates how different VQG questions are from VQA and CQA questions.

In Figure 3 we provide statistics for the various annotations on that portion of the MS COCO images which are represented in the $VQG_{coco-5000}$ dataset. In Figure 3(a) we compare the percentage of object-mentions in each of the annotations. Object-mentions are words associated with the gold-annotated object boundary boxes[5] as provided with the MS COCO dataset. Naturally, COCO captions (green bars) have the highest percentage of these literal objects. Since object-mentions are often the answer to VQA and CQA questions, those questions naturally contain objects less frequently. Hence, we see that VQG questions include the mention of more of those literal objects. Figure 3(b) shows that COCO captions have a larger vocabulary size, which reflects their longer and more descriptive sentences. VQG shows a relatively large vocabulary size as well, indicating greater diversity in question formulation than VQA and CQA. Moreover, Figure 3(c) shows that the verb part of speech is represented with high frequency in our dataset.

[4] http://mscoco.org/

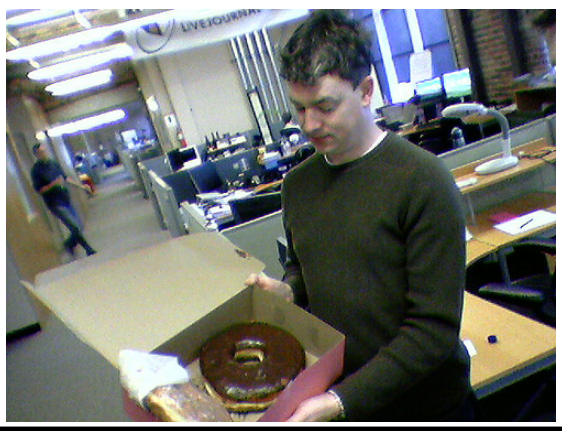

| Dataset | Annotations |
|---|---|
| COCO | - A man holding a box with a large chocolate covered donut. |
| CQA | - What is the man holding with a large chocolate-covered doughnut in it? |
| VQA | - Is this a large doughnut? |
| **VQG** | - Why is the donut so large?<br>- Is that for a specific celebration?<br>- Have you ever eaten a donut that large before?<br>- Is that a big donut or a cake?<br>- Where did you get that? |

Table 1: Dataset annotations on the above image.

Figure 3(d) depicts the percentage of abstract terms such as 'think' or 'win' in the vocabulary. Following Ferraro et al. (2015), we use a list of most common abstract terms in English (Vanderwende et al., 2015), and count all the other words except a set of function words as concrete. This figure supports our expectation that VQG covers more abstract concepts. Furthermore, Figure 3(e) shows inter-annotation textual similarity according to the BLEU metric (Papineni et al., 2002). Interestingly, VQG shows the highest inter-annotator textual similarity, which reflects on the existence of consensus among human for asking a natural question, even for object-centric images like the ones in MS COCO.

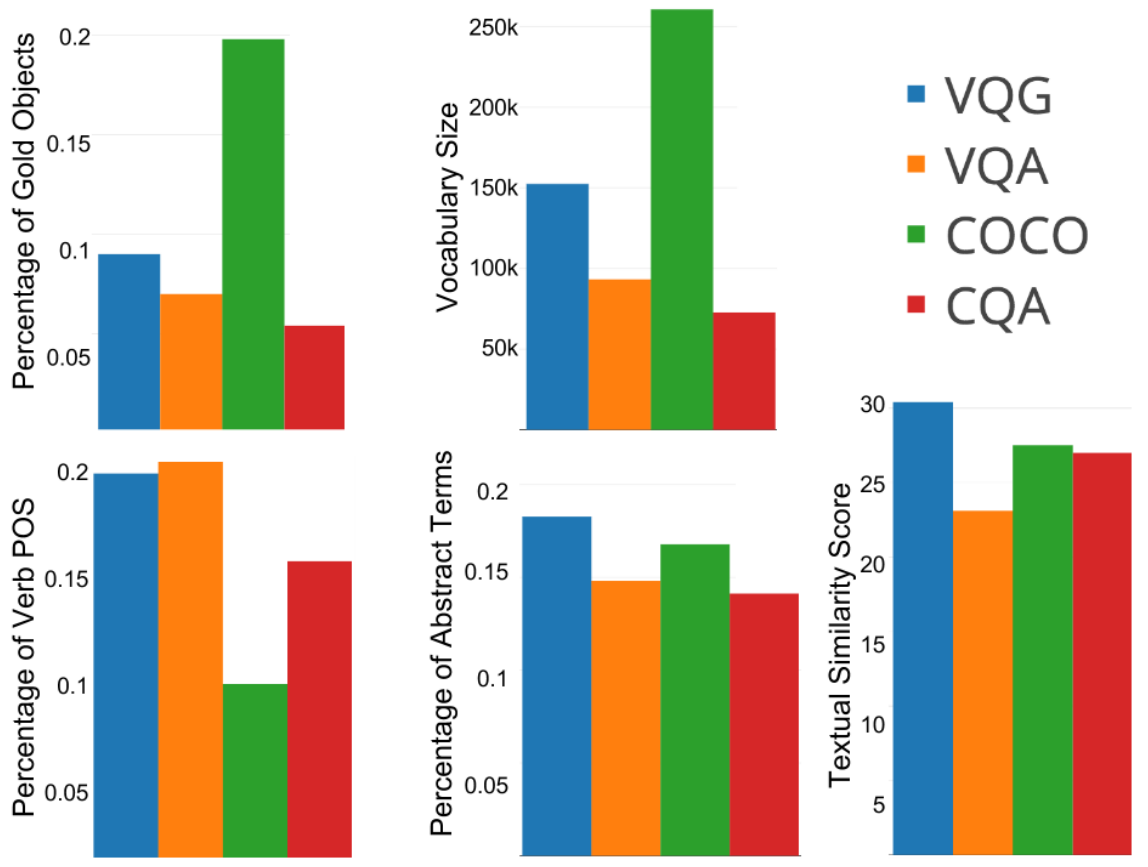

Figure 3: Comparison of various annotations on the MS COCO dataset. (a) Percentage of gold objects used in annotations. (b) Vocabulary size (c) Percentage of verb POS (d) Percentage of abstract terms (e) Inter-annotation textual similarity score.

The MS COCO dataset is limited in terms of the concepts it covers, due to its pre-specified set of object categories. Word frequency in $VQG_{coco-5000}$ dataset, as demonstrated in Figure 4, bears this out, with the words 'cat' and 'dog' the fourth and fifth most frequent words in the dataset. Not shown in the frequency graph is that words such as 'wedding', 'injured', or 'accident' are at the very bottom of frequency ranking list. This observation motivated the collection of the $VQG_{Flickr-5000}$ dataset, with images appearing as the middle photo in a story-full photo album (Huang et al., 2016) on Flickr[6]. The details about this dataset can be found in the supplementary material.

## 3.2 $VQG_{Bing-5000}$

To obtain a more representative visualization of specific event types, we queried a search engine[7] with 1,200 event-centric query terms which were obtained as follows: we aggregated all 'event' and 'process' hyponyms in WordNet (Miller, 1995), 1,000 most frequent TimeBank events (Pustejovsky et al., 2003) and a set of manually curated 30 stereotypical events, from which we selected the top 1,200 queries based on Project Gutenberg word frequencies. For each query, we collected the first four to five images retrieved, for a total of 5,000 images, having first used crowdsourcing to filter out images depicting graphics and cartoons. A similar word frequency analysis shows that the $VQG_{Bing-5000}$ dataset indeed contains more words asking about events: *happen*, *work*, *cause* appear in top 40 words, which was our aim in creating the Bing dataset.

[5]Note that MS COCO annotates only 91 object categories.

[6]http://www.flickr.com

[7]https://datamarket.azure.com/dataset/bing/search

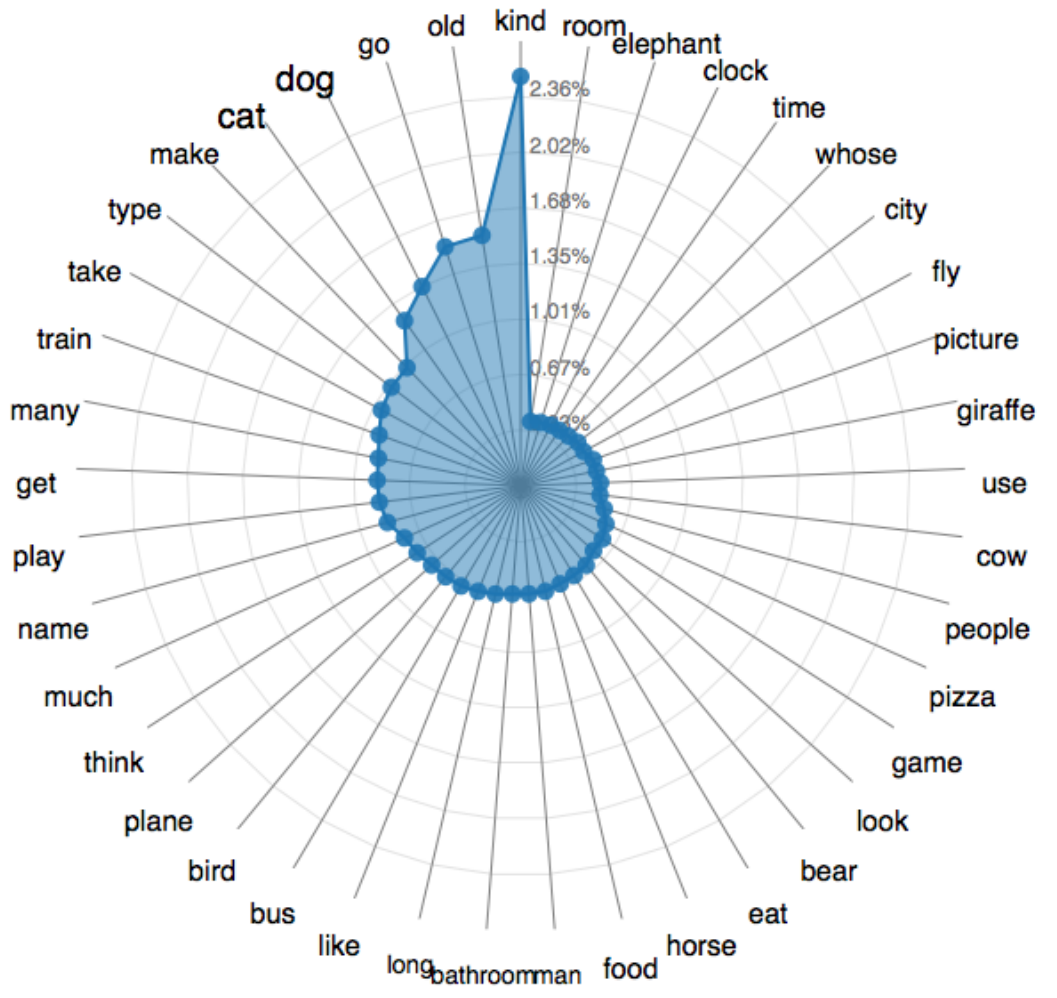

Figure 4: Frequency graph of top 40 words in $VQG_{coco-5000}$ dataset.

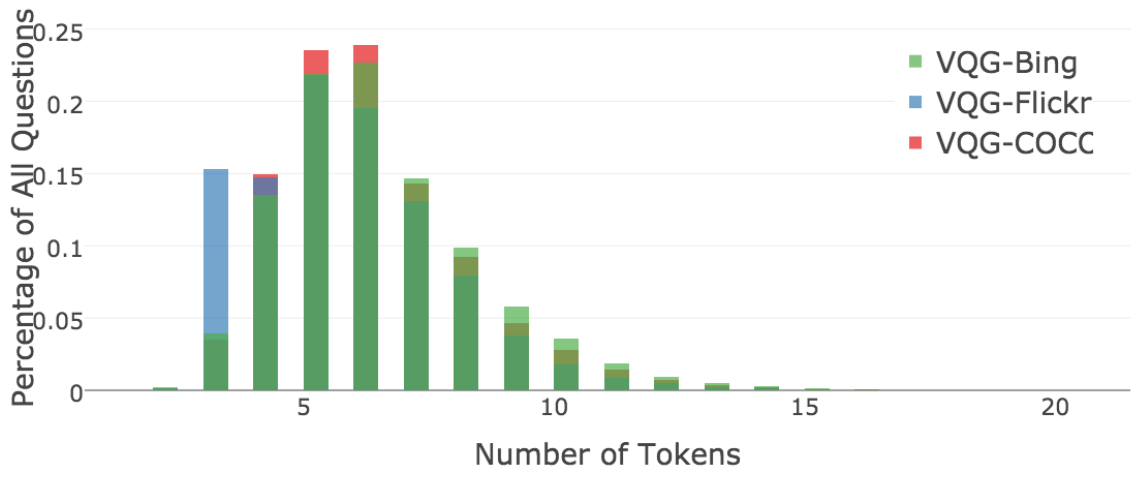

Figure 5: Average annotation length of the three VQG datasets.

**Statistics:** Our three datasets together cover a wide range of visual concepts and events, totaling 15,000 images with 75,000 questions. Figure 5 draws the histogram of number of tokens in VQG questions, where the average question length is 6 tokens. Figure 6 visualizes the n-gram distribution (with n=6) of questions in the three VQG datasets[8]. Table 2 shows the statistics of the crowdsourcing task.

### 3.3 $Captions_{Bing-5000}$

The word frequencies of questions about the $\boldsymbol{VQG_{Bing-5000}}$ dataset indicate that this dataset is substantially different from the MS COCO dataset. Human evaluation results of a recent work (Tran et al., 2016) further confirms the significant image captioning quality degradation on out-of-domain data. To further explore this difference, we crowdsourced 5 captions for each image in the $VQG_{Bing-5000}$ dataset using the same prompt as used to source the MS COCO captions. We call this new dataset $\boldsymbol{Captions_{Bing-5000}}$. Table 3 shows the results of testing the state-of-the-art MSR captioning system on the $\boldsymbol{Captions_{Bing-5000}}$ dataset as compared to the MS COCO dataset, measured by the standard BLEU (Papineni et al., 2002) and METEOR (Denkowski and Lavie, 2014) metrics. The wide gap in the results further confirms that indeed the $\boldsymbol{VQG_{Bing-5000}}$ dataset covers a new class of images; we hope the availability of this new dataset will encourage including more diverse domains for image captioning.

[8]Please refer to our web page on `http://research.microsoft.com/en-us/downloads` to get a link to a dynamic visualization and statistics of all n-gram sequences.

| | |
|---|---|
| # all images | **15,000** |
| # questions per image | **5** |
| # all workers participated | 308 |
| Max # questions written by one worker | 6,368 |
| Average work time per worker (sec) | 106.5 |
| Median work time per worker (sec) | 23.0 |
| Average payment per question (cents) | 6.0 |

Table 2: Statistics of crowdsourcing task, aggregating all three datasets.

| **BLEU** | | **METEOR** | |
|---|---|---|---|
| *Bing* | *MS COCO* | *Bing* | *MS COCO* |
| 0.101 | 0.291 | 0.151 | 0.247 |

Table 3: Image captioning results

Together with this paper we are releasing an extended set of VQG dataset to the community. We hope that the availability of this dataset will encourage the research community to tackle more end-goal oriented vision & language tasks.

## 4 Models

In this Section we present several generative and retrieval models for tackling the task of VQG. For all the forthcoming models we use the VGGNet (Simonyan and Zisserman, 2014) architecture for computing deep convolutional image features. We primarily use the 4096-dimensional output the last fully connected layer ($fc7$) as the input to the generative models.

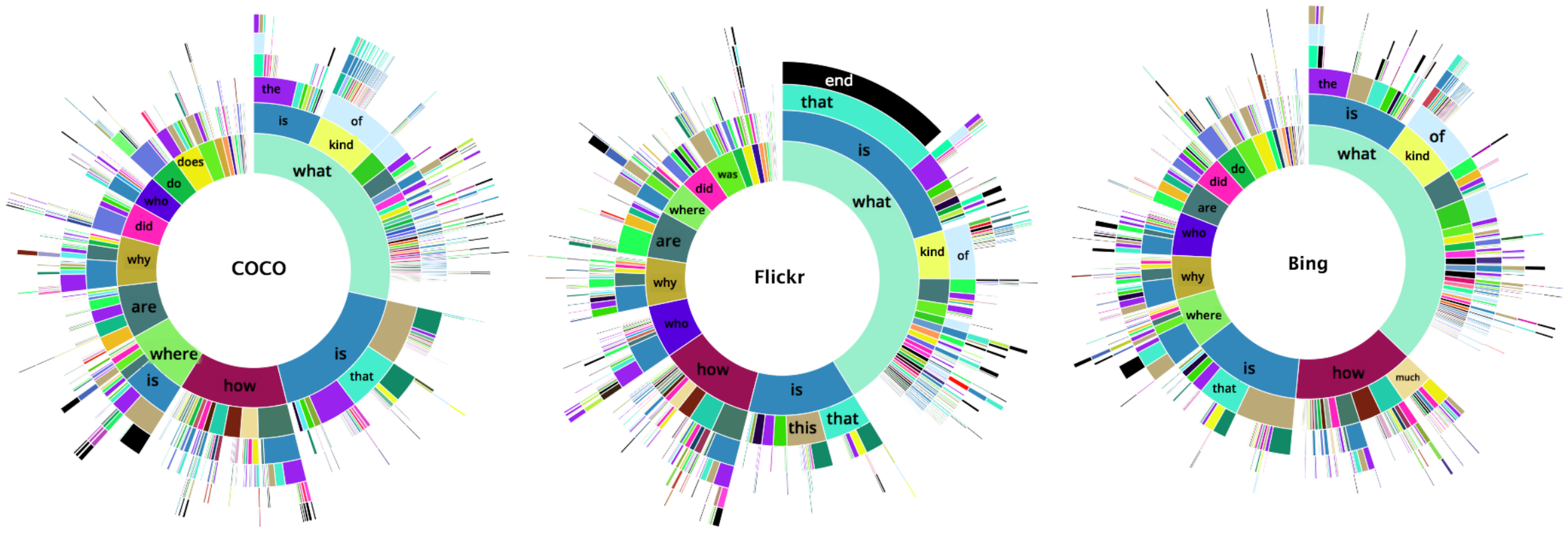

Figure 6: VQG N-gram sequences. 'End' token distinguishes natural ending with n-gram cut-off.

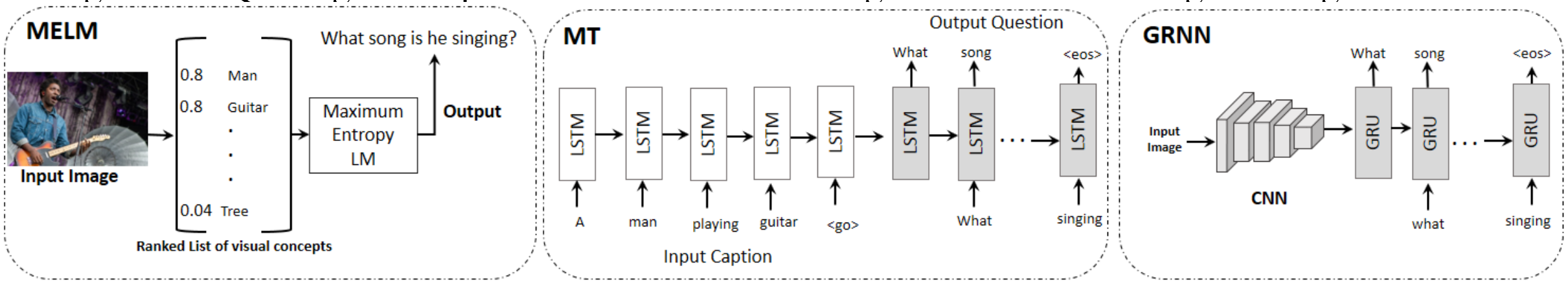

Figure 7: Three different generative models for tackling the task of VQG.

### 4.1 Generative Models

Figure 7 represents an overview of our three generative models. The MELM model (Fang et al., 2014) is a pipeline starting from a set of candidate word probabilities which are directly trained on images, which then goes through a maximum entropy (ME) language model. The MT model is a Sequence2Sequence translation model (Cho et al., 2014; Sutskever et al., 2014) which directly translates a description of an image into a question, where we used the MS COCO captions and $Captions_{Bing-5000}$ as the source of translation. These two models tended to generate less coherent sentences, details of which can be found in the supplementary material. We obtained the best results by using an end-to-end neural model, GRNN, as follows.

**Gated Recurrent Neural Network (GRNN)**: This generation model is based on the state-of-the-art multimodal Recurrent Neural Network model used for image captioning (Devlin et al., 2015; Vinyals et al., 2015). First, we transform the $fc7$ vector to a 500-dimensional vector which serves as the initial recurrent state to a 500-dimensional Gated Recurrent Unit (GRU). We produce the output question one word at a time using the GRU, until we hit the end-of-sentence token. We train the GRU and the transformation matrix jointly, but we do not back-propagate the CNN due to the size of the training data. The neural network is trained using Stochastic Gradient Descent with early stopping, and decoded using a beam search of size 8. The vocabulary consists of all words seen 3 or more times in the training, which amounts to 1942 unique tokens in the full training set. Unknown words are mapped to to an <unk> token during training, but we do not allow the decoder to produce this token at test time.

### 4.2 Retrieval Methods

Retrieval models use the caption of a nearest neighbor training image to label the test image (Hodosh et al., 2013; Devlin et al., 2015; Farhadi et al., 2010; Ordonez et al., 2011). For the task of image captioning, it has been shown that up to 80% of the captions generated at test time by a near state-of-the-art generation approach (Vinyals et al., 2015) were exactly identical to the training set captions, which suggests that reusing training annotations can achieve good results. Moreover, basic nearest neighbor approaches to image captioning on the MS COCO dataset are shown to outperform generation models according to automatic metrics (Devlin et al., 2015). The performance of retrieval models of course depends on the diversity of the dataset.

We implemented several retrieval models customized for the task of VQG. As the first step, we compute K nearest neighbor images for each test image using the $fc7$ features to get a candidate pool. We obtained the most competitive results by setting K dynamically, as opposed to the earlier works which fix K throughout the testing. We observed that candidate images beyond a certain distance made the pool noisy, hence, we establish a parameter called *max-distance* which is an upper bound for including a neighbor image in the pool. Moreover, our experiments showed that if there exists a very similar image to the test image, the candidate pool can be ignored and that test image should become the only candidate[9]. For addressing this, we set a *min-distance* parameter. All these parameters were tuned on the corresponding validation sets using the Smoothed-BLEU (Lin and Och, 2004) metric against the human reference questions.

| | 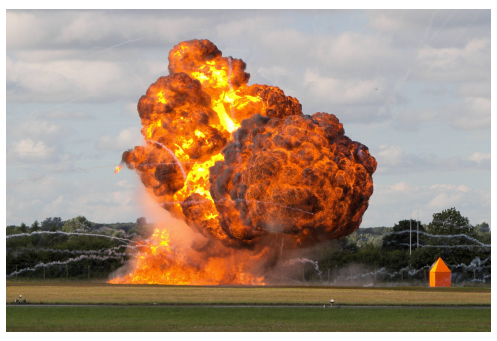 | 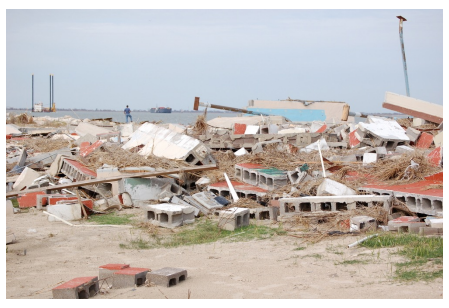 | 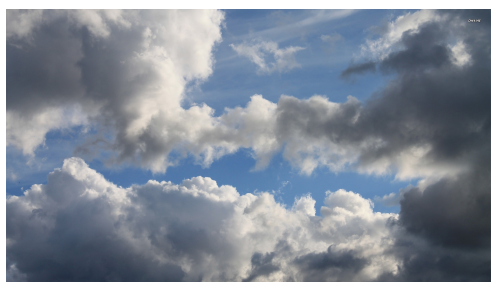 | 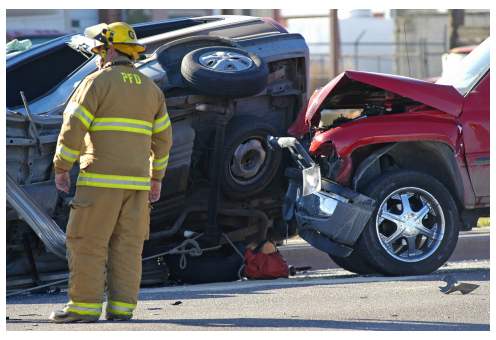 |
|---|---|---|---|---|
| **Q.** | Explosion | Hurricane | Rain Cloud | Car Accident |
| **Human** | - What caused this explosion?<br>- Was this explosion an accident? | - What caused the damage to this city?<br>- What happened to this place? | - Are those rain clouds?<br>- Did it rain? | - Did the drivers of this accident live through it?<br>- How fast were they going? |
| **GRNN** | - How much did the fire cost?<br>- What is being burned here? | - What happened to the city?<br>- What caused the fall? | - What kind of clouds are these?<br>- Was there a bad storm? | - How did the car crash?<br>- What happened to the trailer? |
| **KNN** | - What caused this fire? | - What state was this earthquake in? | - Did it rain? | - Was anybody hurt in this accident? |
| **Caption** | - A train with smoke coming from it. | - A pile of dirt. | - Some clouds in a cloudy day. | - A man standing next to a motorcycle. |

Table 4: Sample generations by different systems on $VQG_{bing-5000}$, in order: Human$_{consensus}$ and Human$_{random}$, GRNN$_{bing}$ and GRNN$_{all}$, KNN+min$_{bleu-all}$, MSR captions. Q is the query-term.

Given that each image in the pool has five questions, we define the one-best question to be the question with the highest semantic similarity[10] to the other four questions. This results in a pool of K candidate questions. The following settings were used for our final retrieval models:

– **1-NN**: Set K=1, which retrieves the closest image and emits its one-best.

– **K-NN+min**: Set K=30 with *max-distance* $= 0.35$, and *min-distance* $= 0.1$. Among the 30 candidate questions (one-best of each image), find the question with the highest similarity to the rest of the pool and emit that: we compute the textual similarity according the two metrics, Smoothed-BLEU and Average-Word2Vec (gensim)[11].

Table 4 shows a few example images along with the generations of our best performing systems. For more examples please refer to the web page of the project.

## 5 Evaluation

While in VQG the set of possible questions is not limited, there is consensus among the natural questions (discussed in Section 3.1) which enables meaningful evaluation. Although human evaluation is the ideal form of evaluation, it is important to find an automatic metric that strongly correlates with human judgment in order to benchmark progress on the task.

### 5.1 Human Evaluation

The quality of the evaluation is in part determined by how the evaluation is presented. For instance,

[9] At test time, the frequency of finding a train set image with *distance* $\leq 0.1$ is 7.68%, 8.4% and 3.0% in COCO, Flickr and Bing datasets respectively.

[10] We use BLEU to compute textual similarity. This process eliminates outlier questions per image.

[11] Average-Word2Vec refers to the sentence-level textual similarity metric where we compute the cosine similarity between two sentences by averaging their word-level Word2Vec (Mikolov et al., 2013) vector representations. Here we use the GenSim software framework (Řehůřek and Sojka, 2010).

it is important for the human judges to see various system hypotheses at the same time in order to give a calibrated rating. We crowdsourced our human evaluation on AMT, asking three crowd workers to each rate the quality of candidate questions on a three-point semantic scale.

### 5.2 Automatic Evaluation

The goal of automatic evaluation is to measure the similarity of system-generated question hypotheses and the crowdsourced question references. To capture n-gram overlap and textual similarity between hypotheses and references, we use standard Machine Translation metrics, BLEU (Papineni et al., 2002) and METEOR (Denkowski and Lavie, 2014). We use BLEU with equal weights up to 4-grams and default setting of METEOR version 1.5. Additionally we use $\Delta$BLEU (Galley et al., 2015) which is specifically tailored towards generation tasks with diverse references, such as conversations. $\Delta$BLEU requires rating per reference, distinguishing between the quality of the references. For this purpose, we crowd-sourced three human ratings (on a scale of 1-3) per reference and used the majority rating.

The pairwise correlational analysis of human and automatic metrics is presented in Table 6, where we report on Pearson's $r$, Spearman's $\rho$ and Kendall's $\tau$ correlation coefficients. As this table reveals, $\Delta$BLEU strongly correlates with human judgment and we suggest it as the main evaluation metric for testing a VQG system. It is important to note that BLEU is also very competitive with $\Delta$BLEU, showing strong correlations with human judgment. Hence, we recommend using BLEU for any further benchmarking and optimization purposes. BLEU can also be used as a proxy for $\Delta$BLEU for evaluation purposes whenever rating per reference are not available.

### 5.3 Results

In this section, we present the human and automatic metric evaluation results of the models introduced earlier. We randomly divided each VQG-5000 dataset into train (50%), val (25%) and test (25%) sets. In order to shed some light on differences between our three datasets, we present the evaluation results separately on each dataset in Table 5. Each model (Section 4.2) is once trained on all train sets, and once trained only on its corresponding train set (represented as X in the results table). For quality control and further insight on the task, we include two human annotations among our models: 'Human$_{consensus}$' (the same as one-best) which indicates the consensus human annotation on the test image and 'Human$_{random}$' which is a randomly chosen annotation among the five human annotations.

It is quite interesting to see that among the human annotations, Human$_{consensus}$ achieves consistently higher scores than Human$_{random}$. This further verifies that there is indeed a common intuition about what is the most natural question to ask about a given image. As the results of human evaluation in Table 5 shows, GRNN$_{all}$ performs the best as compared with all the other models in 2/3 of runs. All the models achieve their best score on $VQG_{COCO-5000}$, which was expected given the less diverse set of images. Using automatic metrics, the GRNN$_X$ model outperforms other models according to all three metrics on the $VQG_{Bing-5000}$ dataset. Among retrieval models, the most competitive is K-NN+min_bleu_all, which performs the best on $VQG_{COCO-5000}$ and $VQG_{Flickr-5000}$ datasets according to BLEU and $\Delta$BLEU score. This further confirms our effective retrieval methodology for including *min-distance* and n-gram overlap similarity measures. Furthermore, the boost from 1-NN to K-NN models is considerable according to both human and automatic metrics. It is important to note that none of the retrieval models beat the GRNN model on the Bing dataset. This additionally shows that our Bing dataset is in fact more demanding, making it a meaningful challenge for the community.

## 6 Discussion

We introduced the novel task of 'Visual Question Generation', where given an image, the system is tasked with asking a natural question. We provide three distinct datasets, each covering a variety of images. The most challenging is the Bing dataset, requiring systems to generate questions with event-centric concepts such as 'cause', 'event', 'happen', etc., from the visual input. Furthermore, we show that our Bing dataset presents challenging images to the state-of-the-art captioning systems. We encourage the community to report their system results on the Bing test dataset and according to the $\Delta$BLEU automatic metric. All the datasets will be released to the public[12].

This work focuses on developing the capabil-ity to ask relevant and to-the-point questions, a key intelligent behavior that an AI system should demonstrate. We believe that VQG is one step towards building such a system, where an engaging question can naturally start a conversation. To continue progress on this task, it is possible to increase the size of the training data, but we also expect to develop models that will learn to generalize to unseen concepts. For instance, consider the examples of system errors in Table 7, where visual features can be enough for detecting the specific set of objects in each image, but the system cannot make sense of the combination of previously unseen concepts. Another natural future extension of this work is to include question generation within a conversational system (Sordoni et al., 2015; Li et al., 2016), where the context and conversation history affect the types of questions being asked.

[12]Please find Visual Question Generation under http://research.microsoft.com/en-us/downloads.

| | | $Human_{consensus}$ | $Human_{random}$ | $GRNN_X$ | $GRNN_{all}$ | $1\text{-}NN_{bleu-X}$ | $1\text{-}NN_{gensim-X}$ | $K\text{-}NN\text{+}min_{bleu-X}$ | $K\text{-}NN\text{+}min_{gensim-X}$ | $1\text{-}NN_{bleu-all}$ | $1\text{-}NN_{gensim-all}$ | $K\text{-}NN\text{+}min_{bleu-all}$ | $K\text{-}NN\text{+}min_{gensim-all}$ |
|---|---|---|---|---|---|---|---|---|---|---|---|---|---|
| | | | | | | | **Human Evaluation** | | | | | | |
| | Bing | 2.49 | 2.38 | 1.35 | **1.76** | 1.72 | 1.72 | 1.69 | 1.57 | 1.72 | 1.73 | 1.75 | 1.58 |
| | COCO | 2.49 | 2.38 | 1.66 | 1.94 | 1.81 | 1.82 | 1.88 | 1.64 | 1.82 | 1.82 | **1.96** | 1.74 |
| | Flickr | 2.34 | 2.26 | 1.24 | **1.57** | 1.44 | 1.44 | 1.54 | 1.28 | 1.46 | 1.46 | 1.52 | 1.30 |
| | | | | | | | **Automatic Evaluation** | | | | | | |
| BLEU | Bing | 87.1 | 83.7 | **12.3** | 11.1 | 9.0 | 9.0 | 11.2 | 7.9 | 9.0 | 9.0 | 11.8 | 7.9 |
| | COCO | 86.0 | 83.5 | 13.9 | 14.2 | 11.0 | 11.0 | 19.1 | 11.5 | 10.7 | 10.7 | **19.2** | 11.2 |
| | Flickr | 84.4 | 83.6 | 9.9 | 9.9 | 7.4 | 7.4 | 10.9 | 5.9 | 7.6 | 7.6 | **11.7** | 5.8 |
| MET. | Bing | 62.2 | 58.8 | **16.2** | 15.8 | 14.7 | 14.7 | 15.4 | 14.7 | 14.7 | 14.7 | 15.5 | 14.7 |
| | COCO | 60.8 | 58.3 | 18.5 | 18.5 | 16.2 | 16.2 | **19.7** | 17,4 | 15.9 | 15.9 | 19.5 | 17.5 |
| | Flickr | 59.9 | 58.6 | 14.3 | **14.9** | 12.3 | 12.3 | 13.6 | 12.6 | 12.6 | 12.6 | 14.6 | 13.0 |
| ΔBLEU | Bing | 63.38 | 57.25 | **11.6** | 10.8 | 8.28 | 8.28 | 10.24 | 7.11 | 8.43 | 8.43 | 11.01 | 7.59 |
| | COCO | 60.81 | 56.79 | 12.45 | 12.46 | 9.85 | 9.85 | 16.14 | 9.96 | 9.78 | 9.78 | **16.29** | 9.96 |
| | Flickr | 62.37 | 57.34 | 9.36 | 9.55 | 6.47 | 6.47 | 9.49 | 5.37 | 6.73 | 6.73 | **9.8** | 5.26 |

Table 5: Results of evaluating various models according to different metrics. X represents training on the corresponding dataset in the row. Human score per model is computed by averaging human score across multiple images, where human score per image is the median rating across the three raters.

| | METEOR | BLEU | ΔBLEU |
|---|---|---|---|
| $r$ | **0.916** (4.8e-27) | 0.915 (4.6e-27) | 0.915 (5.8e-27) |
| $\rho$ | 0.628 (1.5e-08) | 0.67 (7.0e-10) | **0.702** (5.0e-11) |
| $\tau$ | 0.476 (1.6e-08) | 0.51 (7.9e-10) | **0.557** (3.5e-11) |

Table 6: Correlations of automatic metrics against human judgments, with p-values in parentheses.

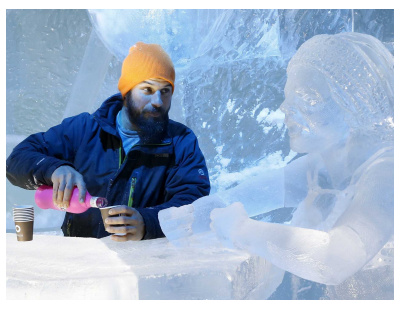

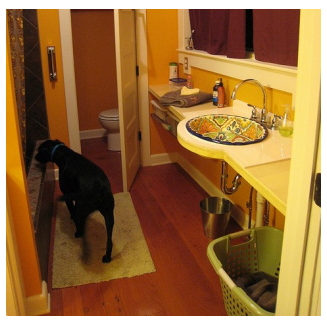

| | | |
|---|---|---|
| **Human** | - How long did it take to make that ice sculpture? | - Is the dog looking to take a shower? |
| **GRNN** | - How long has he been hiking? | - Is this in a hotel room? |
| **KNN** | - How deep was the snow? | - Do you enjoy the light in this bathroom? |

Table 7: Examples of errors in generation. The rows are $Human_{consensus}$, $GRNN_{all}$, and $KNN\text{+}min_{bleu-all}$.

## Acknowledgment

We would like to thank the anonymous reviewers for their invaluable comments. We thank Larry Zitnick and Devi Parikh for their helpful discussions regarding this work, Rebecca Hanson for her great help in data collection, Michel Galley for his guidelines on evaluation, and Bill Dolan for his valuable feedback throughout this work.